# Federated Learning based on Pruning and Recovery

1st Chengjie Ma
Beijing University of Posts and Telecommunications
Beijing Key Laboratory of Intelligent Telecommunication Software and Multimedia
Beijing, China
macj163@163.com

*Abstract*—A novel federated learning training framework for heterogeneous environments is presented, taking into account the diverse network speeds of clients in realistic settings. This framework integrates asynchronous learning algorithms and pruning techniques, effectively addressing the inefficiencies of traditional federated learning algorithms in scenarios involving heterogeneous devices, as well as tackling the staleness issue and inadequate training of certain clients in asynchronous algorithms. Through the incremental restoration of model size during training, the framework expedites model training while preserving model accuracy. Furthermore, enhancements to the federated learning aggregation process are introduced, incorporating a buffering mechanism to enable asynchronous federated learning to operate akin to synchronous learning. Additionally, optimizations in the process of the server transmitting the global model to clients reduce communication overhead. Our experiments across various datasets demonstrate that: (i) significant reductions in training time and improvements in convergence accuracy are achieved compared to conventional asynchronous FL and HeteroFL; (ii) the advantages of our approach are more pronounced in scenarios with heterogeneous clients and non-IID client data.

## I. INTRODUCTION

Existing FL algorithms such as FedAVG, SCAFFOLD[1] and FedOpt [2] rely on the assumption that each participating client has similar resources and can execute the same model locally. While existing gradient compression [3] or model pruning techniques[4] can reduce costs at the expense of a small loss of accuracy, these studies only consider homogeneous client scenarios, where it is assumed that all participating clients have similar computational power and can run on the same model architecture with the same reduced model complexity. However, in most real-world applications, the computational resources of different clients often differ significantly[23, 24]. This heterogeneity leads to insufficient client resources to participate in certain FL tasks that require large models.

Various drawbacks arise when applying classical FL to resource-constrained devices[7]:(1) Unreliability of heterogeneous devices. Heterogeneous devices in the server are more likely to be unable to continue participating in training due to various unexpected conditions. (2) Inefficient rounds. Because of the differences in the device resources of different clients, different clients have different training times per round, and faster devices need to wait for the local models uploaded by slower devices in synchronized federated learning. (3) Low data utilization. Since nodes will tend to choose devices with efficient algorithms and ignore data from poorly performing clients to overcome the above challenges, Asynchronous Federated Learning (AFL), in which the central server performs global model aggregation immediately after collecting local models, is a scheme to try. Asynchronous FL schemes[8], [9], [10]update the global model whenever the server receives a local model from each device; in particular, in FedAsyn, the global model is updated asynchronously based on the device's staleness, i.e., the time difference between the current round time when the device first receives the global model and the previous round when it received the global model. Asynchronous schemes are very effective in dealing with dropouts, and FedFix [11] extends the standard aggregation scheme so that asynchronous federated learning can maintain the convergence stability of synchronous aggregation.

Another way to cope with the performance differences of heterogeneous clients in federated learning is to assign different-sized models to clients with different performance. A more complete model is assigned to clients with stronger performance, and a smaller client is assigned to clients with weaker performance to compensate for the difference in training speed between clients with different performance. HeteroFL[5] proposes to split the global model into smaller models of different sizes along the width depending on the performance of the clients, which preserves the full depth of the DNN architecture for each client and only adjusts the heterogeneous width splitting ratio of the clients. However, this tends to result in very fine and deep sub-networks, which leads to a significant loss of essential features[12], and thus to a drastic degradation of model quality. FedDF [6] proposes to apply integrated distillation to fuse models with different architectures. However, FedDF requires additional datasets for distillation operations and incurs significant overhead between training rounds. ScaleFL[13] adaptively scales the global model along the broadband and depth based on the client's computational resources, and similar work has been done with DepthFL[14]. FedMP [15]uses structured pruning by utilizing a multi-armed slot machine algorithm dynamically determines the pruning rate of different clients based on the resource situation of each node, and proposes an aggregation strategy based on model residuals. S. Liu[16] uses a combination of pruning, client selection, and wireless resource allocation. First, the proposed federated pruning algorithm is analyzed for convergence; then the optimization problem of maximizing the convergence rate in finite time by dynamically controlling the pruning rate, client-side selection, and wireless resource allocation is proposed; finally, an analytical solution for the optimal pruning rate and resource allocation is derived, and a threshold-based client-selection algorithm is proposed to further improve the learning efficiency. S. Vahidian[17] used both structured pruning and unstructured pruning strategies to assign different sizes of models to clients with different performances, and proposed Sub-FedAvg, which improves the aggregation algorithm for federated learning by federating and averaging only the parts that are common to the models of different clients.

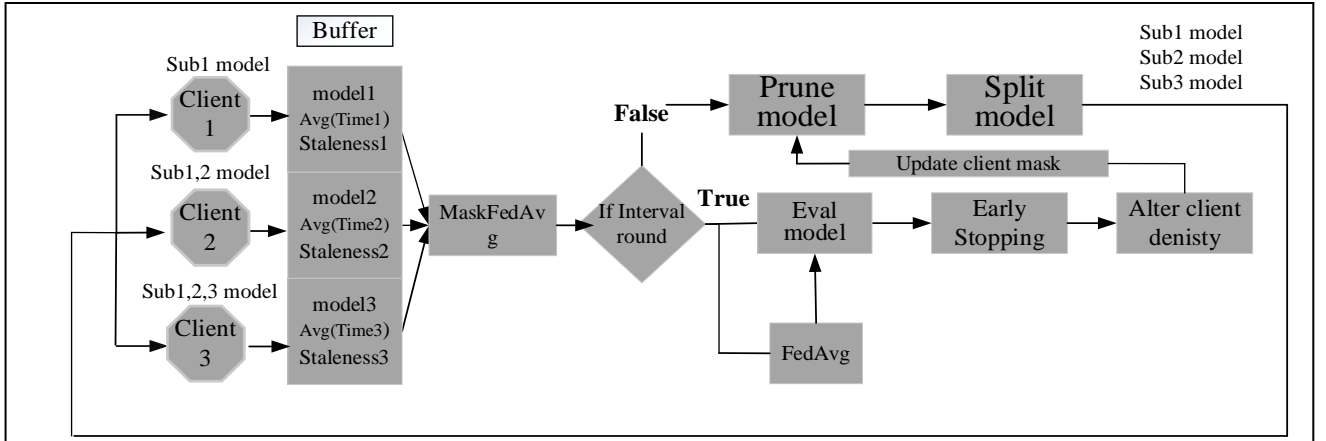

Fig. 1. The overall structure of the algorithm for PR-FL

The main idea of the Federated Learning based on Pruning and Recovery algorithm is to assign smaller local models to clients with limited resources. And an asynchronous approach is used to avoid assigning over-pruned models to clients with extremely limited resources, which may lead to a decrease in the accuracy of the global model. Then at a later stage of model training, the pruned model can be recovered on resource-constrained clients to improve the overall accuracy of the global model. And for this algorithm for federation learning the aggregation process is improved and a buffer machine is used to allow asynchronous federation learning to be trained like synchronous federation learning. The process of sending the global model from the server to the client is also improved to reduce the amount of communication.

The overall structure of the algorithm for PR-FL is shown in Figure 1.

## II. BACKGROUND

### A. Synchronized Federal Learning Time Stream Analysis

**Network buffer**: Considering the network difference between the server and the client in real environment and the download congestion of the server caused by a large number of clients sending local models to the server at the same time, a network buffer is set up. When the uploading speed of the model sender is larger than the downloading speed of the model receiver, the model sender will store the unreceived part of the model receiver in the network buffer first, and when the model receiver downloads all the files in the network buffer, the model transmission process is completed.

That is, in Figure 2, client 1 uploads the local model $W_1^0$ to the server, the client uploads all the local models at the time $t_{u_1}^0$, and the server receives all the models $W_1^0$ at the time $t_{u_1}^{0'}$, $\Delta t_u = t_{u_1}^{0'} - t_{u_1}^0$ depending on the real-time download speed of the downloader and the real-time performance of the uploader. When the performance of the server is strong enough and the number of clients is not too large, the delay of the server in transmitting the global model to the client depends on the download speed of the client, and the delay of the client in transmitting the local model to the server depends on the upload speed of the client.

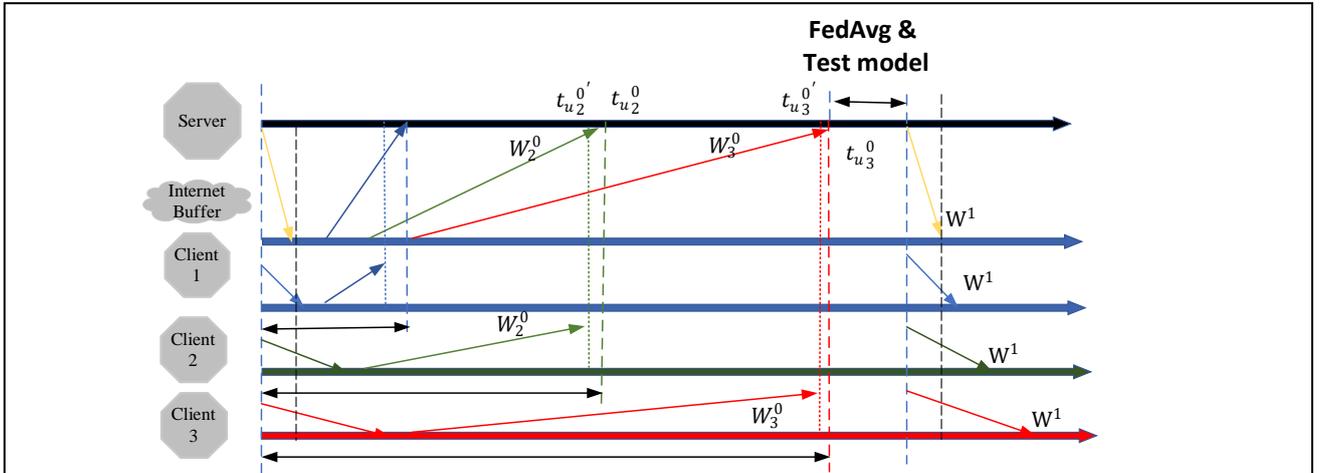

Fig. 2. Synchronized Federated Learning Time Stream Analysis

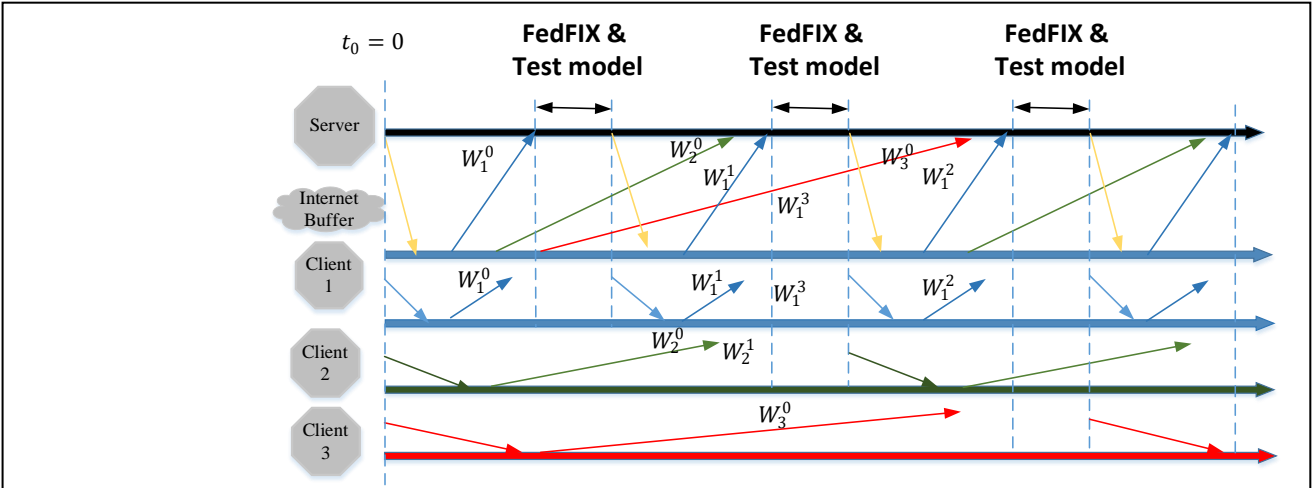

Fig. 3. Asynchronized Federated Learning FedFix Time Stream Analysis

*B. Synchronized Federal Learning Time Stream Analysis*

In the heterogeneous federated learning environment shown in Fig. 3, the process of asynchronous federated learning is illustrated when there are three clients. (1) In the process shown, client 1 is updated roughly twice as often as client 2 and roughly three times as often as client 3. However, this is not static and may change as network fluctuations and differences in arrival times between different clients accumulate. (2) Before the second server aggregation client 2's round 0 model arrives at the server first, but the server continues to wait for a shorter period of time for client 1's round 1 model to arrive at the server as well before performing the server aggregation. This is because when the server's performance is not much higher than the client's performance, if the aggregation is performed for every client model received, the server needs to send global models to the client frequently, which may lead to a new performance bottleneck.

Because the time required for different clients to complete a round of work in a heterogeneous environment varies greatly, the training time of synchronous federated learning is mainly affected by the worst-performing client of the network, and the higher-performing client spends most of its time waiting for the lower-performing client to complete the model transfer. It has a short-board effect. In order to solve the problem of low training efficiency in heterogeneous federated environments, Asynchronous Federated Learning[9] (FedAsyn) is proposed. For Asynchronous Federated Learning, the server no longer waits for all client models to arrive before aggregating them, by setting a time threshold $\Delta t^n$, $T^n = T^{n-1} + \Delta t^n + T_{merge}$, for each round, where $T_{merge}$ is the time needed for the server to receive the models and perform aggregation and other operations, which $T_{merge}$ can be regarded as a small constant due to the high performance of the server. If for the arrival time of the client model $T_i^n > T^n$, the client's local

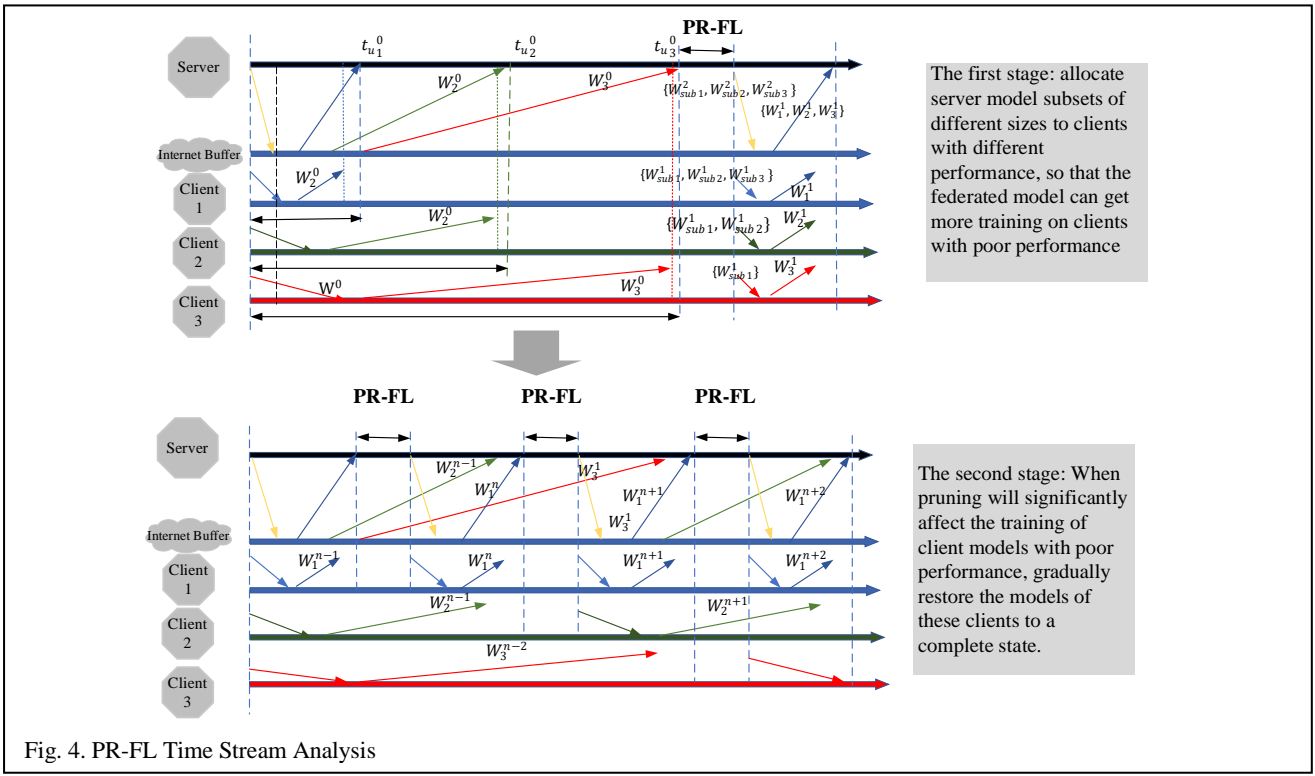

Fig. 4. PR-FL Time Stream Analysis

model cannot participate in the server aggregation process. Set the coefficient as $\rho_i(n)$:

$$\rho_i^j(n) = \begin{cases} 1, & T^{n-1} < T_i^j \leq T^n \\ 0, & else \end{cases}, j \leq n \quad (1)$$

$\rho_i^j(n)$ is whether the client's local model of the first round is received in the first round for creating a new global model. It can be noted here that when = 1 (at this point), and when, there exists a stale model of the client's past rounds received by the server, i.e., the problem of staleness as proposed in FedAsyn[9]. For synchronized federated learning, each round is designed such that it always holds $\rho_i^n(n) = 1$, all clients' local models participate in server aggregation at each round.

Traditional asynchronous federated learning has two main issues, namely (1) Update Staleness: Update staleness causes the server to potentially use outdated client data when aggregating the model. This delay can be quantified through the use of "staleness" parameters. The "freshness" of updates impacts the quality of the global model. If an update is based on a global model calculated a long time ago, it may push the global model in a direction that is no longer optimal. High latency can lead to the convergence of the global model to suboptimal solutions, especially when the data distribution is highly uneven. (2) Unbalanced Training: Some clients may undergo more updates than others, resulting in the model favoring the data distribution of these clients, especially in cases where the data distribution is already non-Independently and Identically Distributed (Non-IID). The model may struggle to generalize well to the data of all clients because it more closely reflects the features of the data from clients that are frequently updated.

III. FEDERATED LEARNING BASED ON PRUNING AND RECOVERY

In this section, we present a topic model specifically designed for unbalanced short text datasets, utilizing co-occurrence word networks. This model addresses the unbalance issue by normalizing both scarce and abundant topics to some extent. It leverages the construction of co-occurrence word networks to effectively capture context information. Furthermore, the model employs techniques inspired by LDA for topic prediction. By combining these approaches, we aim to enhance the performance of topic modeling on unbalanced short text datasets.

A. PR-FL Time Stream Analysis

First Stage: Adjusting the pruning ratio of heterogeneous workers based on the working time of each client in each round, ensuring that different heterogeneous workers obtain training models of different sizes according to their performance differences. This allows heterogeneous clients to undergo asynchronous training at similar training speeds (with low staleness in asynchronous training). This process ensures that clients with lower performance receive sufficient training.

Second Stage: However, after training reaches a certain point, the accuracy loss caused by pruning severely hinders the training process. Gradually, the pruned models are restored to ensure that the final training accuracy of the model can recover to a higher level.

```
Algorithm 1: Algorithm for the PR-FL
1  Initialize: Create a series of early_stopping()
   mechanisms for each client and server model.
2  Set T1=maxT₁ = max_i T_{i,1}, and T_n − T_{n−1} =
   Δt_n  (n > 1) T_n − T_{n−1} = Δt_n  (n >
   1) while T < T_max do
      // Server Procedure:
3     if T = T_n then
4        Update buffer B_n with received models
         w^n_{i,n'(i)}
5        W^n_mask ← MaskFedAvg(B_n)
6        if I divides n + 1 then
7           for each client i do
8              acc_i = test_acc(w^n_{i,n'(i)}, D_test)
9              if early_stopping(acc_i) then
10                if d^n_{i,n'(i)} ≤ 1 then
11                   ρ_{min,i} = min(ρ_i + Δρ, 1)
12                end
13             end
14          end
15          W^n ← FedAvg(B_n)
16          acc = test_acc(W_n, D_test)
17          if early_stopping(acc) then
18             if all d^n_{i,n'(i)} ≥ 1 then
19                break
20             end
21          end
22          Calculate each ρ_i for client i using Q_i:
             t̄^n_i = (1/|Q_i|) Σ_{k=1}^{|Q_i|} t_k
             ρ_i = min(min_j t̄^n_j / t̄^n_i, ρ_{max,i})
23          MASK^n ← Prune_model(W^n_mask, P)
24       end
         // Send model to clients
25       ΔW^n ← Split_model(W^n_mask, MASK^n)
26    end
      // Parallel Client Procedure:
27    while client receives the new model W^n_mask do
28       Merge the ΔW^n to obtain the local model
         w_{i,n}
29       Client trains the local model with local data:
         w_{i,n+1} ← w_{i,n} − η∇L(w_{i,n})
30       Client i sends w^{n+1}_i back to the server
31    end
32 end
```

As shown in Figure 4: In the first stage, pruning techniques are used to allocate different sizes of model subsets to clients based on their network performance. This makes the training time per round more evenly distributed among different clients, alleviating the issues of update staleness and unbalanced training brought about by traditional asynchronous federated learning. However, using pruning techniques on the model can compromise its accuracy. When the model reaches a certain stage of training, the accuracy loss from pruning becomes more severe than the issues of update staleness and unbalanced training. At this point, the complete model from clients with better performance is gradually used to restore the models of clients with poorer performance. This ensures that federated learning continues to train at a faster pace. In the later stages of federated training, this algorithm is similar to traditional asynchronous federated learning, but the training paradigm presented in this paper significantly mitigates the issues of update staleness and unbalanced training in asynchronous federated learning.

*B. The overall algorithmic process of PR-FL*

Define the client set $C = \{c_1, c_2, ..., c_M\}$. Client $i$ in the $n$-th round has a model on the server's buffer denoted as $w_{i,n'(i)}^n$. This model is the one sent by the server in the $n'(i)$-th round($0 \leq n'(i) \leq n$), and its staleness is $s_i^n$. The time taken by client $i$ to complete the training in the previous round is $t_{i,n'(i)}^n$, which includes network transmission time and local model training time. Let $Q_i$ be a first-in-first-out (FIFO) queue structure with a maximum length of $|Q_i| \leq I$ where $I$ is the server's pruning interval. It stores the most recent $I$ times $t_{i,n'(i)}^n$ taken by client $i$ to complete the previous round of training. The density of the client's model is $\rho_{i,n'(i)}^n$.

The client's model buffer is denoted as $B^n = \{b_1^n, \cdots, b_M^n\}$, where $b_i^n = w_{i,n'(i)}^n, i, n'(i), s_i^n, t_{i,n'(i)}^n, Q_i, \rho_{i,n'(i)}^n$. To ensure model accuracy, the minimum density of a model is $\rho_{min}$, and the minimum density for clients is $P_{min} = \{\rho_{min,1}, \rho_{min,2}, ..., \rho_{min,M}\}$. The growth increment of model density is denoted as $\Delta\rho$, typically set to 0.2.

Firstly, in the first stage, to ensure that poorly performing clients receive sufficient training, smaller models are assigned to these clients. Since the primary time delay for client participation in federated training is communication time, and client communication volume decreases as the model size decreases, the density of each client is initially determined based on the average communication time over the client's previous period. Specifically:

$$\rho_i = max\left(\frac{\min_j \overline{t_i^n}}{\overline{t_i^n}}, \rho_{min,i}\right) \quad (2)$$

$\overline{t_i^n}$ is the average time taken by client $i$ over the last $|Q_i|$ rounds, and $\min_j \overline{t_i^n}$ is the minimum average time taken by any client over the last $|Q_i|$ rounds, This formula aims to allow clients to train smaller models while ensuring that their density is not lower than the specified minimum. The goal is to encourage resource limited clients to participate actively in federated training, enabling the global model to learn more from clients with poorer performance.

$$\overline{t_i^n} = \frac{1}{|Q_i|}\sum_{k=1}^{|Q_i|} t_k \quad (3)$$

$$\rho_i = max\left(\frac{\min_j \overline{t_i^n}}{\overline{t_i^n}}, \rho_{min,i}\right) \quad (4)$$

$\overline{t_i^n}$ is the average time taken by client $i$ over the last $|Q_i|$ rounds, and $\min_j \overline{t_i^n}$ is the minimum average time taken by any client over the last $|Q_i|$ rounds, This formula aims to allow clients to train smaller models while ensuring that their density is not lower than the specified minimum. The goal is to encourage resource limited clients to participate actively in federated training, enabling the global model to learn more from clients with poorer performance.

However, in the later stages of the federated learning training phase, due to the significant accuracy loss caused by pruning hindering the training process, there is a gradual recovery of pruned models. To gauge when the accuracy loss from model pruning becomes a significant hindrance to the training process, this paper employs the early_stopping() mechanism for evaluation. Specifically, when the client's model accuracy has not shown gradual improvement over multiple training rounds, indicating that the model is no longer effectively learning additional knowledge, it becomes necessary to gradually recover the model. This is achieved by increasing $\rho_{min,i}$ to ensure that the client receives models from denser clients. In other words:

$$\rho_{min,i} = min(\rho_{i,n'(i)}^n + \Delta\rho, 1) \quad (5)$$

The federated learning training process concludes when the density of all clients is set to 1, and the model's accuracy no longer exhibits gradual improvement with training.

*C. MaskFedAvg model aggregation algorithm*

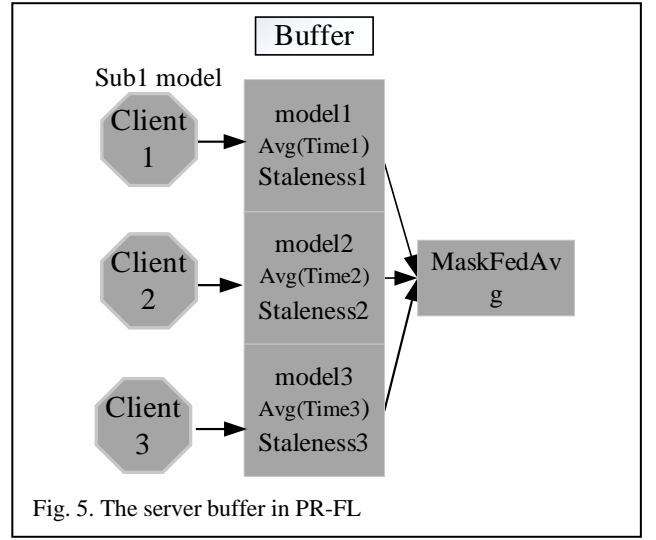

Fig. 5. The server buffer in PR-FL

In traditional asynchronous federated learning, the aggregation process involves proportionally combining the global model on the server with newly arrived models based on their staleness. However, this method faces instability issues when dealing with client models arriving at short intervals, as the cumulative multiplication approach cannot guarantee model stability.

In PR-FL, during the model aggregation process, the server first establishes a buffer for each client, storing the latest models of each client. Whenever a new client model is received by the server, it updates the buffer and performs complete aggregation based on the stored client's latest model and staleness.

Subsequently, this method performs model aggregation in two ways, obtaining the global FedAvg model and the global MaskFedAvg model. These models are used for model accuracy testing and splitting into new client sub-models, respectively. Since this method allocates different-sized sub-models to different clients, directly applying the FedAvg algorithm would lead to averaging of neurons, including pruned neurons, causing some weights to rapidly shrink during the aggregation process.

To ensure that pruned neurons are not averaged multiple times, this method employs the MaskFedAvg approach to aggregate models from the buffer, obtaining the global MaskFedAvg model distributed to each client. Then, the method utilizes the FedAvg approach to aggregate the models from the buffer once again, resulting in the global FedAvg model used to test the accuracy of the global model. Experiments indicate that the testing results of the global FedAvg model are relatively more stable compared to the global MaskFedAvg model.

Select an appropriate method to calculate $s_i(n)$, and then the MaskFedAvg process is shown in Figure 4-4. The calculation is as follows: to ensure that there is no significant variation in the data range of neurons in the model, use $mask_i^n$ to represent the mask vector received by the server for client $i$'s model $W_{i,n'}^n$. In other words, if the corresponding part in $W_{i,n'}^n$ is 0, it indicates pruning, and if it is 1, it indicates non-pruning.

$$(mask_i^n)_{a,b,c\cdots} = \begin{cases} 1, & if\ (w_{i,n'(i)}^n)_{a,b,c\cdots} \neq 0 \\ 0, & if\ (w_{i,n'(i)}^n)_{a,b,c\cdots} = 0 \end{cases} \quad (6)$$

Calculate the aggregation weight $p_i^n$ for each client based on $s_i^n$, according to the following formula:

$$p_i^n = \frac{s_i^n}{\sum s_i^n} \quad (7)$$

Calculate the cumulative weight sum $W_{acc}$ and the cumulative mask sum $mask_{acc}$ for each client separately.

$$W_{acc} = \sum_{i=0}^{m} p_i^n \times w_{i,n'(i)}^n \quad (8)$$

$$mask_{acc} = \sum_{i=0}^{m} p_i^n \times mask_i^n \quad (9)$$

Subsequently, perform element-wise division on $W_{acc}$ and $mask_{acc}$. This yields the change in the current model. However, to prevent potential elements from being zero after aggregation, which could lead to actual pruning of the server-side model and render it irrecoverable, it is necessary to restore portions with zero values in the change of the new server model using the $W_{mask}^{n-1}$ from the old server model, as follows:

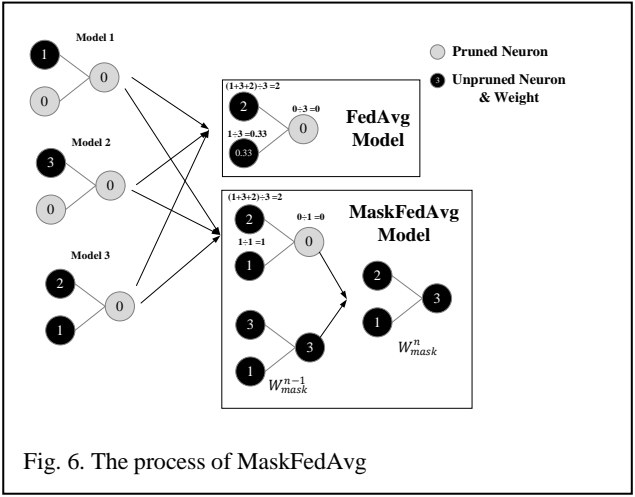

Fig. 6. The process of MaskFedAvg

**Algorithm 2: MaskFedAvg Model Aggregation**

**Input:** the previous MaskFedAvg model $W_{mask}^{n-1}$ the server buffer $B^n$
**Output:** the aggregated masked model $W_{mask}^n$ for the current round
// Server Procedure:
1 Initialize an accumulator for the models weight and masks: $W_{acc} \leftarrow 0$, $Mask_{acc} \leftarrow 0$
2 **for** each client model $w_{i,n'(i)}^n$ **do**
3     Compute the Aggregate weight $\rho_i^n$ for the current client's model: $\rho_i^n \leftarrow \frac{s_i^n}{\sum s_i^n}$
4     Accumulate the weighted model: $W_{acc} \leftarrow W_{acc} + w_{i,n'(i)}^n \cdot \rho_i^n$
5     Computer the mask for the client's model: $(mask_i^n)_{a,b,c,\dots} = \begin{cases} 1, & if\ (w_{i,n'(i)}^n)_{a,b,c,\dots} \neq 0 \\ 0, & if\ (w_{i,n'(i)}^n)_{a,b,c,\dots} = 0 \end{cases}$
6     Accumulate the weighted mask: $Mask_{acc} \leftarrow Mask_{acc} + \rho_i^n \cdot mask_i^n$
7 **end**
8 **for** each element $a, b, c, \dots$ **do**
9     **if** $(Mask_{acc}^n)_{a,b,c,\dots} \neq 0$ **then**
10       $(W_{mask})_{a,b,c,\dots} \leftarrow \frac{(W_{acc})_{a,b,c,\dots}}{(Mask_{acc})_{a,b,c,\dots}}$
11     **else**
12       $(W_{mask}^n)_{a,b,c,\dots} \leftarrow (W_{mask}^{n-1})_{a,b,c,\dots}$
13     **end**
14 **end**
15 Update the server model with the global learning rate $\eta_g$: $W_{mask}^n \leftarrow (1-\eta_g) \cdot W_{mask}^{n-1} + \eta_g \cdot W_{mask}^n$

$$(W_{mask}^n)_{local} = \begin{cases} (W_{mask}^{n-1})_{a,b,c\cdots}, & if\ (mask_{acc})_{a,b,c\cdots} = 0 \\ \frac{(W_{acc})_{a,b,c\cdots}}{(mask_{acc})_{a,b,c\cdots}}, & if\ (mask_{acc})_{a,b,c\cdots} \neq 0 \end{cases} \quad (10)$$

Considering the global learning rate $\eta_g$ on the server side, then:

$$W_{mask}^n = (1-\eta_g)W_{mask}^{n-1} + \eta_g \times (W_{mask}^n)_{local} \quad (11)$$

$W_{mask}^n$ is the server model used for model splitting and sent to each client in the $n$-th round.

### D. Differential Model Distribution

.During the process of sending the model from the server to the clients, as different clients receive models of varying sizes, distributing models of different sizes directly to each client can lead to redundant and duplicated models being sent

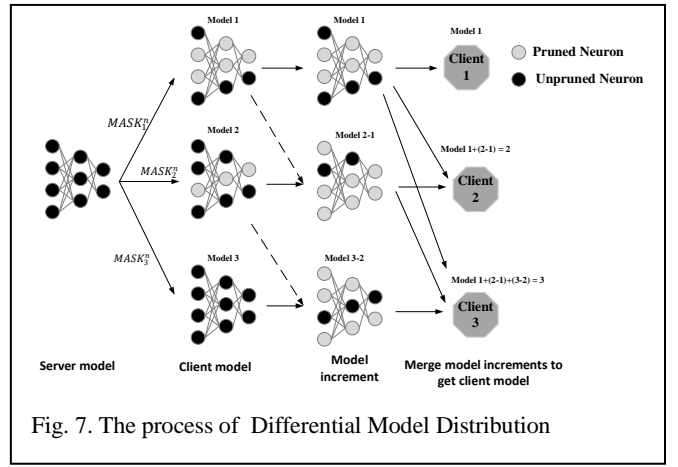

Fig. 7. The process of Differential Model Distribution

```
Algorithm 3: Model Split Algorithm
Input: MaskFedAvg model W_mask^n and the
        corresponding mask set MASK^n
Output: submodel set ΔW^n for w_i^n and the
        corresponding index list Index^n
// Server Procedure:
1 Initialize W_0^n as the zero matrix matching the
  dimensions of client models;
2 for each client C_i do
3   | Determine the required model subset for C_i
      using MASK^n and assign it as
      w_i^n = W_mask^n ⊙ MASK_i^n;
4 end
// Note: W_i^n is ordered by
   ascending index i
5 for each client model W_i^n, starting with i = 1 do
6   | Compute the weight
      increment: Δw_i^n = w_i^n − w_(i−1)^n;
7   | Upload Δw_i^n in the network buffer for access by
      client; Construct the index range
      Index_i^n = [1, i] and transmit Index_i^n to C_i;
8 end
// Parallel Client Procedure:
9 Receive Index_i^n from the server;
10 for each submodel indicated by Index_i^n do
11   | if incremental updates are needed then
12   |   | Download the corresponding ΔW_i^n from the
            server;
13   |   | Merge ΔW_i^n to update the local model and
            obtain the full W_i^n;
14   | end
15 end
```

from the server. This becomes one of the bottlenecks in the federated learning training process, known as the model upload bottleneck. To address this issue, a new model distribution paradigm is proposed.

The server initially sorts the sub-models to be sent to each client in ascending order of size. For ease of representation, let's assume that the sizes of the sub-models sent to each client increase from small to large, i.e., $\rho(w_i^n) > (w_{(i-1)}^n)$, $\rho(w_i^n))$, where $\rho(w_i^n)$ represents the density of model $w_i^n$. The differential model is then obtained through the following equation:

$$\Delta w_i^n = w_i^n - w_{(i-1)}^n \quad (12)$$

$\Delta w_i^n$ represents the subset of neurons that constitute the increased portion of the larger sub-model $w_i^n$ relative to $w_{(i-1)}^n$. This is then utilized through the following equation:

$$Index_i^n = [1, i] \quad (13)$$

$[1, i]$ denotes the sequence from 1 to $i$. Clients determine the differential model subset they need, $w_i^n$, through $Index_i^n$. Initially, the server sends the client the indices of the differential models it needs to receive. The server starts by uploading the smallest-sized client model and subsequently sends increments of other client models relative to the previous model. This approach eliminates the need for the server to send redundant parts of different client models. Upon receiving the model indices from the server, clients sequentially download the required sub-models. Once all the necessary models are downloaded, synthesizing all the sub-models results in obtaining the client model for this iteration.

As shown in Figure 6, suppose the server needs to send models to three clients with varying performance, requiring models with densities of 1, 0.66, and 0.33, respectively. In the traditional model transmission approach, the server uploads each of the three models separately, resulting in a total transmission volume of approximately 1.99 times the server model size. However, with this distribution method, the server first sends the client the indices of the models it needs to receive. The server then starts by uploading the smallest-sized client model, followed by sending increments of other client models relative to the previous model. This way, the server avoids sending redundant parts of different client models. For instance, in the first transmission, it sends the sub-model with the smallest density of 0.33. In the second transmission, it sends the increment of the 0.66 model minus the 0.33 model, requiring only the transmission of the increment with a density of 0.33. Similarly, in the third transmission, it only sends the increment of the 1.0 model relative to the 0.66 increment model. Clients, upon receiving the model indices from the server, sequentially download the required sub-models. Once all the necessary models are downloaded, synthesizing all the sub-models results in obtaining the client model for this iteration. This method results in a total server transmission volume of 1 times the server model size, and it does not significantly increase with the number of clients and sub-models.

IV. EXPERIMENTS

A. Settings

The federated learning progressive pruning technique was evaluated on four image classification tasks as follows:

(a) Conv-2 model on FEMNIST [18].

(b) VGG-11 model [19] on CIFAR-10 [20].

For the FEMNIST experiment, data from 193 authors corresponding to FEMNIST were selected. The VGG-11 model was adjusted for this experiment, modifying some layers to match the output label count in the dataset. The number of clients was set to 10, and for FEMNIST, all images from the 193 authors were distributed among the clients (the first 9 clients each having images from 19 authors, and the last one having images from 22 authors).

Baselines. We compare the test accuracy vs. time curve of PR-FL with four baselines: FedAsyn, FedFIX[11], FedAvg[21] and HeteroFL. Except for the method itself, all other configurations remain consistent with PR-FL.

The parameters for federated learning were configured as shown in TABLE I.

TABLE I. PARAMETERS FOR FEDERATED LEARNING

| Dataset | CIFAR-10/FEMNIST |
|---|---|
| Local Training | $0.25 \cdot 0.5^{\frac{r}{10000}}/0.25$ |
| Batchsize | 20 |
| Local Training Iterations | 5 |
| Pruning Interval | 50 |
| Total Rounds | 20000/10000 |

TABLE III. THE TRAINING ROUNDS FOR DIFFERENT CLIENTS.

| Client ID | 1 | 2 | 3 | 4 | 5 | 6 | 7 | 8 | 9 | 10 |
|---|---|---|---|---|---|---|---|---|---|---|
| FedAsyn | 10000 | 9940 | 9878 | 9548 | 5484 | 4517 | 2822 | 2460 | 2453 | 1999 |
| NoRecover-PR-FL | 10000 | 9951 | 9936 | 9986 | 9935 | 9923 | 9995 | 9952 | 9968 | 9997 |

TABLE IV. THE TIME REQUIRED TO ACHIEVE THE CORRESPONDING ACCURACY.

| | Cifar10 | | | | Femnist | | | |
|---|---|---|---|---|---|---|---|---|
| | 80% | | 84.0% | | 80.0% | | 83.5% | |
| | IID | NO-IID | IID | NO-IID | IID | NO-IID | IID | NO-IID |
| PR-FL | 26300 | 29700 | 39100 | 52500 | 10700 | 12400 | 23400 | 26900 |
| HeteroFL | 40900 | 44500 | 70900 | 97900 | 17700 | 23200 | 40900 | 41600 |
| FedFix | 48000 | 59100 | 108700 | 111500 | 20300 | 25700 | 43900 | 57400 |
| FedAsyn | 70100 | 82800 | 113700 | 172500 | 39300 | 40100 | 61200 | 81600 |
| FedAvg | 202200 | 210900 | - | - | 136300 | 136700 | - | - |

TABLE V. ACCURACY ACHIEVED AT MAXIMUM TIME.

| | Accuracy at 25,000 seconds on CIFAR-10 | | Accuracy at 14,000 seconds on FEMNIST | |
|---|---|---|---|---|
| | IID | NO-IID | IID | NO-IID |
| PR-FL | 87.12 | 86.85 | 85.97 | 85.21 |
| HeteroFL | 86.61 | 84.72 | 84.59 | 84.32 |
| FedFix | 86.65 | 85.43 | 85.20 | 84.44` |
| FedAsyn | 86.34 | 85.41 | 84.13 | 83.26 |
| FedAvg | 82.34 | 81.91 | 79.22 | 79.84 |

Due to the scenario involving compression optimization in a real-world setting on heterogeneous devices, it is necessary to simulate the passage of time using a fixed time interval for realistic scenario emulation. In this context, the network transmission speeds for each client are simulated using a log-exponential distribution with a mean of 0 and a variance of 0.3, while the server's network speed is modeled using a log-exponential distribution with a mean of 0 and a variance of 0.1. The logarithmic exponential results are then multiplied by the corresponding network speeds to simulate the fluctuation of actual network environments. Fig. 7. illustrates log-exponential distributions with different variances.

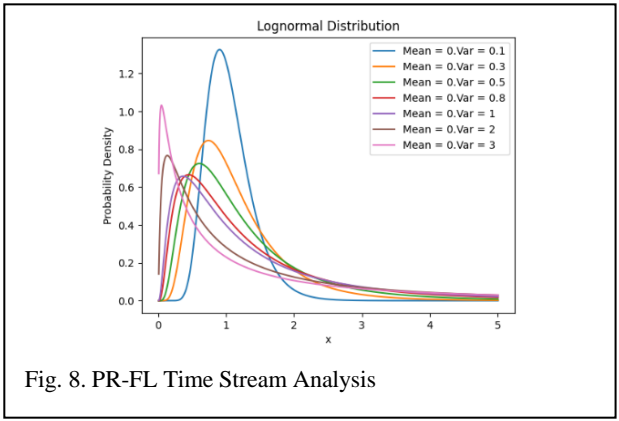

Fig. 8. PR-FL Time Stream Analysis

In different datasets, the network environment settings for each client are outlined in TABLE II. These settings include download and upload speeds for both the server and clients, as well as the complete model size and minimum model density. As observed in [21], upload speeds are generally significantly lower than download speeds.

TABLE II. NETWORK TRANSMISSION SETTINGS

| Dataset | CIFAR-10/ FEMINIST |
|---|---|
| Server Download Speed | 100 |
| Server Upload Speed | 20 |
| Client Upload Speeds | 5.0, 4.0, 3.0, 2.5, 1.5, 1.0, 0.6, 0.5, 0.5, 0.4 |
| Client Download Speeds | 20, 18, 12, 10, 6, 4, 2.5, 2.0, 2.0, 1.5 |
| Complete Model Size (MB) | 39.20/26.5 |
| Minimum Model Density | 0.1/0.05 |

These settings provide a more realistic environment for evaluating compression optimization in federated learning on simulated heterogeneous devices.

### B. Client Network Quality and Training Time Analysis

In order to compare the relationship between different client network qualities, model sizes, and the number of training rounds for each client, this experiment selected the FEMNIST dataset along with its corresponding network environment. The FedAsyn method directly employs asynchronous techniques in the model. NoRecover-PR-FL is a weakened version of PR-FL that does not perform model recovery. Instead, it prunes clients based on their network quality. For instance, if a client's network performance is half that of the fastest network, it is allocated 50% of the model for training.

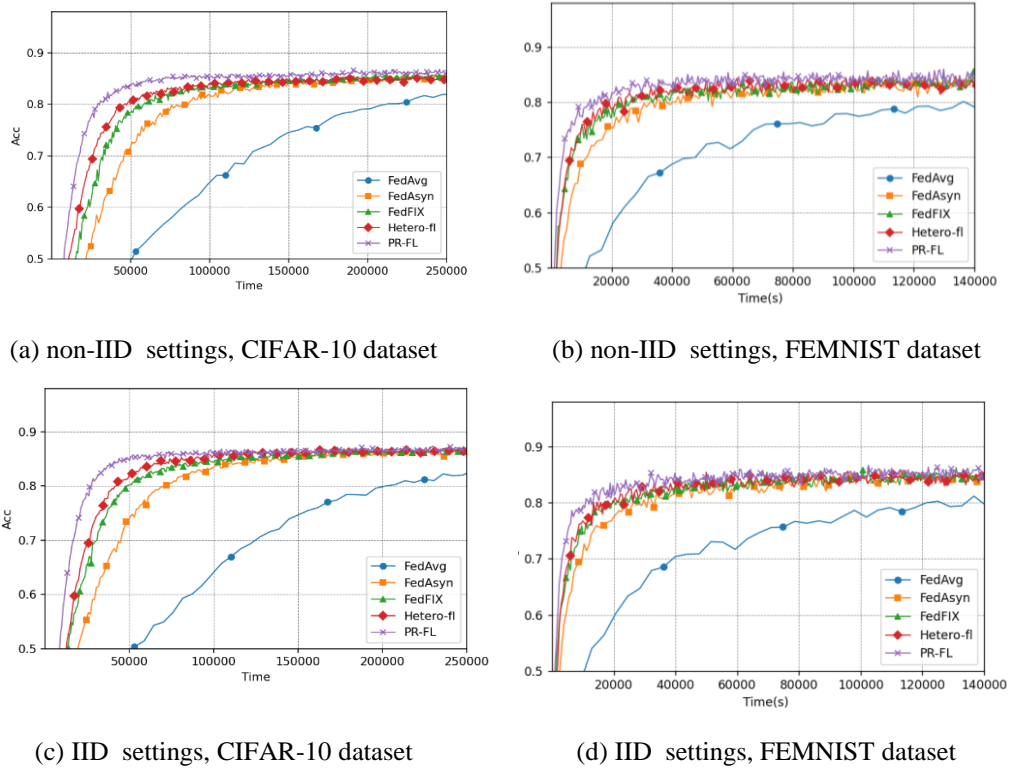

(a) non-IID settings, CIFAR-10 dataset

(b) non-IID settings, FEMNIST dataset

(c) IID settings, CIFAR-10 dataset

(d) IID settings, FEMNIST dataset

Fig. 9. the relationship between the test accuracy and time for PR-FL and other benchmark algorithms in a heterogeneous environment

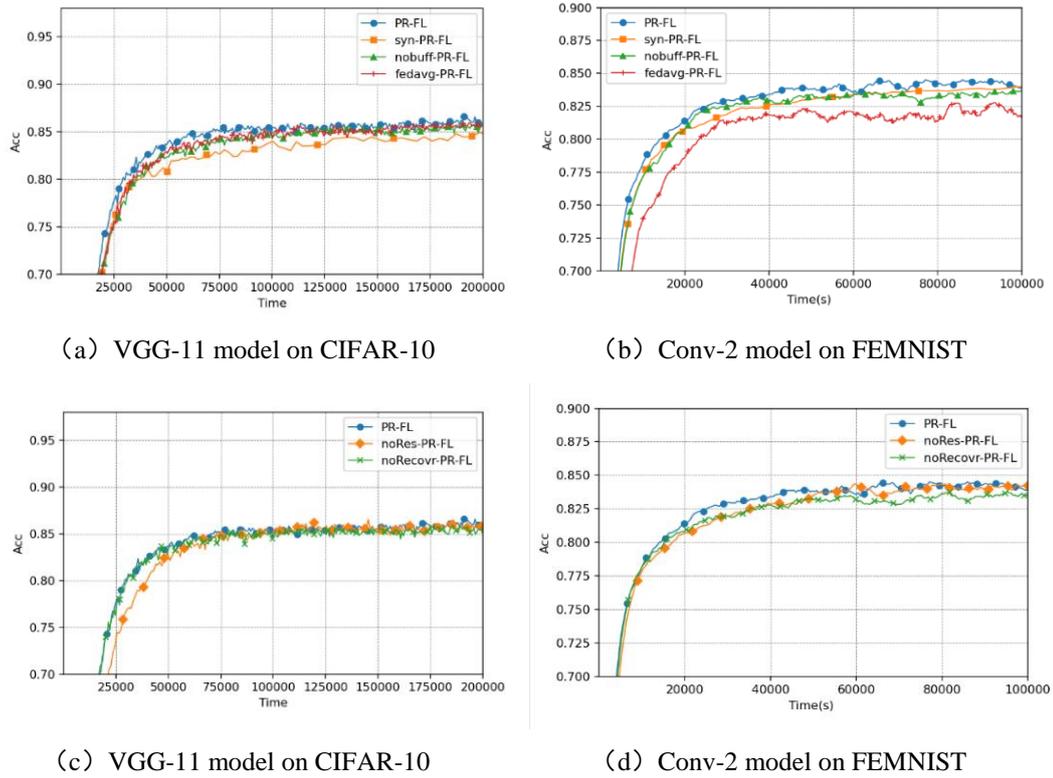

(a) VGG-11 model on CIFAR-10

(b) Conv-2 model on FEMNIST

(c) VGG-11 model on CIFAR-10

(d) Conv-2 model on FEMNIST

Fig. 10. The relationship between the accuracy and time for PR-FL and sub-models in a heterogeneous environment

TABLE III. reveals that in an asynchronous environment, if client models are not pruned, clients with poorer network performance undergo fewer training rounds. However, this relationship is not strictly proportional. The server's performance is not sufficiently robust, leading to instances where some clients experience waiting due to network congestion on the server. For clients with weaker performance, waiting might also occur, but it constitutes a smaller

TABLE I. THE TRAINING ROUNDS FOR DIFFERENT CLIENTS.

| Client ID | 1 | 2 | 3 | 4 | 5 | 6 | 7 | 8 | 9 | 10 |
|---|---|---|---|---|---|---|---|---|---|---|
| FedAsyn | 10000 | 9940 | 9878 | 9548 | 5484 | 4517 | 2822 | 2460 | 2453 | 1999 |
| NoRecover-PR-FL | 10000 | 9951 | 9936 | 9986 | 9935 | 9923 | 9995 | 9952 | 9968 | 9997 |

TABLE VI. THE TIME REQUIRED TO ACHIEVE THE CORRESPONDING ACCURACY.

|  | Cifar10 | | | | Femnist | | | |
|---|---|---|---|---|---|---|---|---|
|  | 80% | | 84.0% | | 80.0% | | 83.5% | |
|  | IID | NO-IID | IID | NO-IID | IID | NO-IID | IID | NO-IID |
| PR-FL | 26300 | 29700 | 39100 | 52500 | 10700 | 12400 | 23400 | 26900 |
| HeteroFL | 40900 | 44500 | 70900 | 97900 | 17700 | 23200 | 40900 | 41600 |
| FedFix | 48000 | 59100 | 108700 | 111500 | 20300 | 25700 | 43900 | 57400 |
| FedAsyn | 70100 | 82800 | 113700 | 172500 | 39300 | 40100 | 61200 | 81600 |
| FedAvg | 202200 | 210900 | - | - | 136300 | 136700 | - | - |

TABLE VII. ACCURACY ACHIEVED AT MAXIMUM TIME.

|  | Accuracy at 25,000 seconds on CIFAR-10 | | Accuracy at 14,000 seconds on FEMNIST | |
|---|---|---|---|---|
|  | IID | NO-IID | IID | NO-IID |
| PR-FL | 87.12 | 86.85 | 85.97 | 85.21 |
| HeteroFL | 86.61 | 84.72 | 84.59 | 84.32 |
| FedFix | 86.65 | 85.43 | 85.20 | 84.44` |
| FedAsyn | 86.34 | 85.41 | 84.13 | 83.26 |
| FedAvg | 82.34 | 81.91 | 79.22 | 79.84 |

proportion as their own network transmission latency plays a more significant role. Clients with stronger performance experience reduced proportions of waiting-related delays due to lower network transmission latency as a percentage of the total delay.

*C. The relationship between accuracy and time.*

As shown in Figure 8, PR-FL by employing model pruning algorithms for clients with lower performance, allows for more training on these clients compared to the FedAsyn and FedFix asynchronous federated learning algorithms. Consequently, this method exhibits a faster training speed compared to asynchronous federated learning algorithms. In contrast to the synchronous federated learning FedAvg method, where a significant portion of training time is wasted waiting for clients with weaker performance, the training speed in a heterogeneous environment is much lower. In the case of HeteroFL, this method incorporates asynchronous settings, reducing the time delay for the model to wait for other clients. Additionally, model recovery is applied in the later stages of training, enabling the federated model to achieve higher accuracy.

*D. The impact of IID datasets on the experiment.*

Fig. 8. presents the experimental results using the dataset independent and identically distributed (IID) setting, revealing a significant reduction in the advantage of PR-FL compared to other asynchronous federated methods in this environment. This is attributed to the phenomenon of client drift, where Non-IID data distribution causes the global model to be more biased towards certain clients' data distributions, preventing the learning of globally optimal information. In traditional asynchronous federated learning, the imbalance in client performance can lead to some clients not receiving sufficient training, exacerbating the issue of client drift. However, PR-FL allows clients with poorer performance to receive more adequate training, alleviating the phenomenon of client drift in the heterogeneous device environment. In environments with IID data distribution, where the client drift phenomenon is already mitigated, the advantage of PR-FL over other asynchronous federated methods is significantly reduced.

*E. Ablation Study Evaluation Metrics*

we evaluated various components of PR-FL. Among them, syn-PR-FL was modified to adhere to the settings of synchronous federated learning; nobuff-PR-FL eliminated the buffering mechanism in PR-FL; fedavg-PR-FL replaced the MaskFedAvg method with the FedAvg method; noRes-PR-FL omitted the model splitting feature, and noRecover-PR-FL skipped model recovery. All other experimental settings remained the same.

As depicted in Fig. 9. (a) and (b), For syn-PR-FL, due to its adherence to synchronous federated learning settings, might result in some clients waiting for others. Additionally, the synchronous waiting issue becomes more severe after model recovery. nobuff-PR-FL, without the buffering method, may lead to more unstable model aggregation and consequently inferior performance. fedavg-PR-FL, when using the FedAvg method for model aggregation, might average some neuron weights with pruned neuron weights, causing certain neurons' overall weights to be excessively reduced and slowing down the training process in federated learning.

In Fig. 9. (c) and (d), sub-models exhibit minimal differences in PR-FL. For noRes-PR-FL, the absence of the differential model splitting setting increases the communication load of the global model from the server to the clients. Regarding noRecover-PR-FL, this setting remains identical to PR-FL in the early stages of model training, differing only in the inability to perform model recovery in the later stages. Experimental results indicate that model recovery contributes to achieving higher accuracy.

## V. CONCLUSION

This paper proposes a novel federated learning training framework in a heterogeneous environment, taking into account the varying network speeds of different clients in real-world scenarios. The framework combines asynchronous learning algorithms with pruning techniques, aiming to address the inefficiencies of traditional federated learning algorithms in heterogeneous device scenarios, as well as the latency and insufficient training of some clients in asynchronous algorithms. Additionally, the paper introduces a progressive model volume recovery during the training process, ensuring both accelerated model training and maintained model accuracy.

Improvements were made to the aggregation process of federated learning in this algorithm, incorporating a buffer mechanism to enable asynchronous federated learning to train similarly to synchronous federated learning. The process of the server sending the global model to clients was also enhanced to reduce communication overhead. The code implementation simulates the communication process between asynchronous federated learning clients and servers using clocks, providing a new perspective for experimental setups in asynchronous federated learning technology.